\documentclass[
twocolumn,
]{ceurart}

\usepackage[utf8]{inputenc}

\hypersetup{breaklinks=true}
\usepackage{color}
\usepackage{xspace}
\usepackage{soul}

\def\MoCo{\emph{MoCo}\xspace}
\def\MAE{\emph{MAE}\xspace}
\def\supViT{\emph{supViT}\xspace}
\def\Swin{\emph{Swin}\xspace}
\def\ResNet18{\emph{ResNet18}\xspace}
\def\DenseNet121{\emph{DenseNet121}\xspace}

\begin{document}


\conference{MRC 2023 – The Fourteenth International Workshop Modelling and Representing Context. Held at ECAI 2023, 30.09.-5.10.2023, Kraków, Poland.}

\title{Does context matter in digital pathology?}



\author[1]{Paulina Tomaszewska}[%
  orcid=0000-0001-6767-1018,
  email=paulina.tomaszewska3.dokt@pw.edu.pl,
]

\author[1]{Mateusz Sperkowski}[%
  email=mateusz.sperkowski.stud@pw.edu.pl,
]

\author[1,2]{Przemysław Biecek}[%
  orcid=0000-0001-8423-1823,
  email=przemyslaw.biecek@pw.edu.pl,
]

\address[1]{Faculty of Mathematics and Information Science, Warsaw University of Technology, Koszykowa 75, 00-662 Warsaw, Poland}
\address[2]{Faculty of Mathematics, Informatics and Mechanics, University of Warsaw, Banacha 2, 02-097 Warsaw, Poland}
\begin{abstract}
The development of Artificial Intelligence for healthcare is of great importance. Models can sometimes achieve even superior performance to human experts, however, they can reason based on spurious features. This is not acceptable to the experts as it is expected that the models catch the valid patterns in the data following domain expertise. In the work, we analyse whether Deep Learning (DL) models for vision follow the histopathologists' practice so that when diagnosing a part of a~lesion, they take into account also the surrounding tissues which serve as context. It turns out that the performance of DL models significantly decreases when the amount of contextual information is limited, therefore contextual information is valuable at prediction time. Moreover, we show that the models sometimes behave in an unstable way as for some images, they change the~predictions many times depending on the size of the context. It may suggest that partial contextual information can be misleading.
\end{abstract}

\begin{keywords}
  digital pathology \sep
  spatial context \sep
  Deep Learning \sep
  Computer Vision
\end{keywords}

\maketitle

\section{Introduction}
Deep Learning (DL) models are perceived as black-boxes. They sometimes make decisions based on different reasons than humans do. DL models tend to take shortcuts and follow spurious correlations~\cite{shortcut}. The popular example is the case where the model misclassified a~husky as a wolf because there was snow in the background, which was a rule in the case of images with wolves within the dataset~\cite{lime}. Although following such a~rule may lead to high classification performance (if the dataset is biased), we expect the model to distinguish between wolves and husky dogs based on the animal features. Motivated by the fact that there is sometimes a mismatch between the~way DL models and humans reason, we decided to investigate whether the DL models for vision follow the~same good practices when diagnosing lesions based on histopathological data as expert histopathologists. The histopathologists when diagnosing a particular region of a lesion, take into account also the surrounding tissue~\cite{random_field}. We~investigate whether the classification performance of DL models for vision will be higher when they have access to information about neighbouring tissues than in the case where no contextual information is given. We conduct a quantitative analysis on how the~amount of contextual information within the input to the~models impacts the final performance. \\
Our contribution is as follows:
\begin{itemize}
    \item we verify whether DL models for vision behave in a similar way to histopathologists who benefit from contextual information when diagnosing lesions 
    \item we measure quantitatively the impact of the~amount of contextual information provided to different DL models for vision on their classification performance 
    \item we investigate whether it happens that the models behave in a non-stable way by changing predictions for the images many times given different amounts of contextual information
\end{itemize}

\begin{figure}[h!]
\includegraphics[width=1\columnwidth]{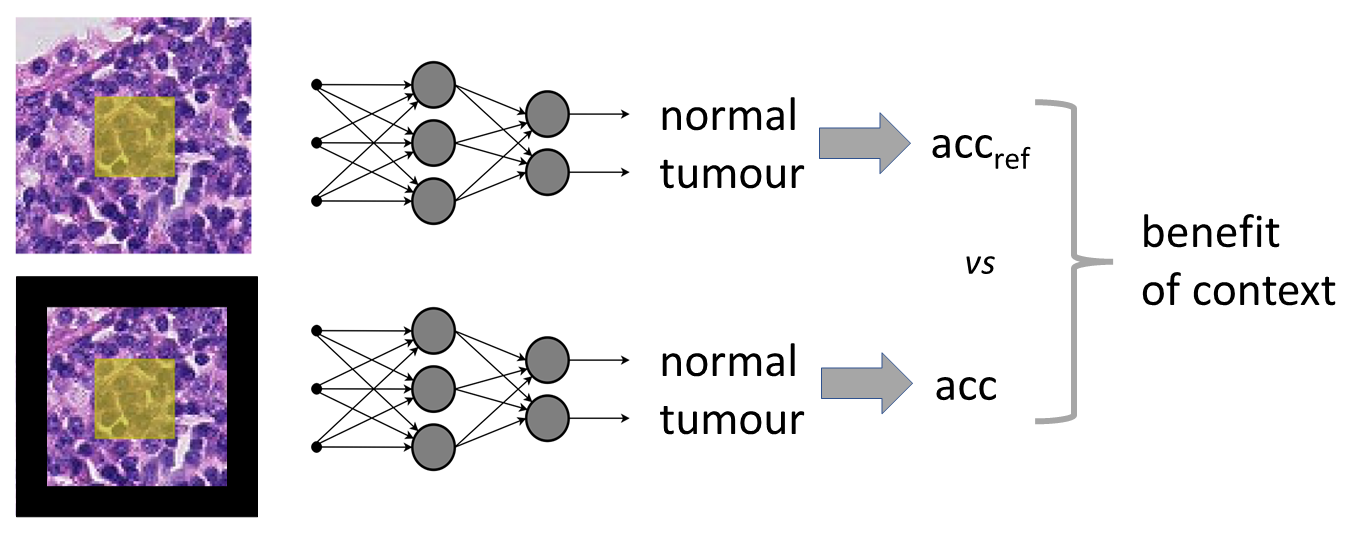}
\caption{The scheme of the proposed study. The yellow squares in the center of histopathological images depict the~regions that the annotations are based on (the squares are shown only for visualization purposes and are not present in dataset images). The black border is applied to remove some parts of contextual information. 
}
\label{fig:scheme}
\end{figure}
The code to replicate our results is available at \texttt{https://github.com/ptomaszewska/PCam\_context}.

\section{Motivation}
The histopathological data is saved in the form of Whole Slide Images (WSIs) where the whole lesion is under huge resolution. In the popular Camelyon16 Breast Cancer dataset~\cite{camelyon} the images have an average resolution of 94,747x188,764 (avg. size of 1.97GB). Such huge images are difficult to load into memory and process, therefore they are most commonly split into smaller-sized patches. The advantage of WSIs is that based on them both local (cell-level) and global (tissue-level) analysis can be performed. In the work, we use the variant of the~Camelyon16 dataset, PatchCamelyon (PCam)~\cite{pcam}, where the~initial WSIs are cut into patches of size 96x96. Each such patch has a label - normal tissue or tumour lesion and the model's task is a binary classification. However, in such a~dataset, the global context is not preserved as the~relationship between neighbouring patches is lost. 
Nevertheless, the dataset contains local context within images since the label is assigned to the whole patch only based on the central 32x32 region of the patch. Therefore, the~surrounding box can be thought of as contextual information. The question is whether contextual information is useful when making predictions.


\section{Method}
\label{sec:method}
The goal of the study is to check whether the DL models for vision are sensitive to different amounts of contextual information within input histopathological images. The method is inspired by the LIME~\cite{lime} which is a permutation-based XAI method where some of the superpixels within images are switched off to check the contribution of each superpixel to the model's prediction.

In our study, first, we use the test set of the PCam dataset as input for inference to the already trained models on the original PCam dataset. The~resulting performance metrics serve as a reference point when the full available context is provided to the model ($acc_{ref}$, $precision_{ref}$, $recall_{ref}$, $AUC_{ref}$).\\
In the following experiments, we restrict the amount of contextual information in the image data.  
Let us define, the size of context ($s$) as a
width in pixels of the area around the central 32x32 square. The~maximum size of the context is $(96-32)/2 = 32$ where 96 is the length of the original image's side. The bigger the context size, the more information about the neighbouring tissue is provided to the model. To~evaluate how the~size of the~context impacts the model prediction, we remove external layers of pixels of the context area and the image is padded to the original size of 96x96 with black pixels (we call it a border for brevity). We decided to use the black colour as it is often used as a baseline colour in explainable AI (XAI) methods i.e. Integrated Gradients~\cite{ig}. We applied the border to the images to obscure some part of the context instead of cutting off the pixels and changing the image resolution to avoid a~situation where it would be difficult to disentangle the source of the~performance change - increased resolution vs. limited context size.\\
The images padded with a~black border are the input to the DL models. We~analyse the difference between the metrics obtained on padded images and the reference, original ones. This will give us information on how much contextual information is beneficial when making predictions.

\begin{figure}[h!]
\centering
\includegraphics[width=0.5\columnwidth]{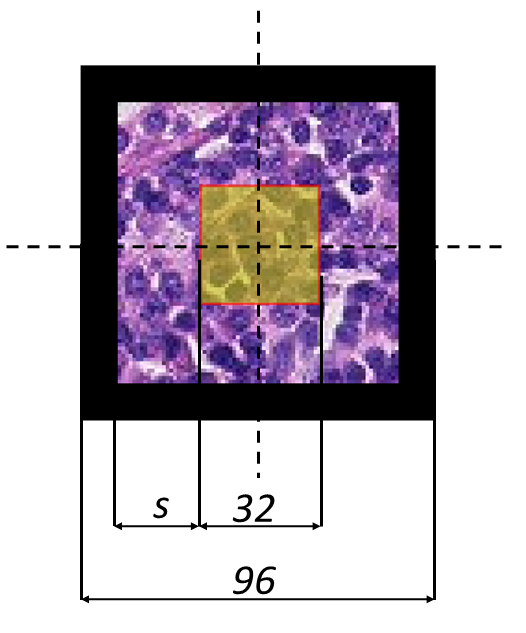}
\caption{The histopathological image padded with black border with dimensions specified. 
 The dimension \textit{s} denotes context size.}
\label{fig:scheme}
\end{figure}

\section{Experiments}
\subsection{Deep Learning models}
We apply the method described in Section~\ref{sec:method} on two classes of DL models - convolutional (\ResNet18 and \DenseNet121) and transformer-based (\Swin~\cite{swin} and \textit{ViT}~\cite{vit}). The convolutional models trained on histopathological data (without any image standarization) were taken from~\cite{toolbox}.
In the case of the transformer-based models, we took the models pretrained on Imagenet. The \Swin model was pretrained in a supervised manner, whereas in the case of \textit{ViT}, we used the models pretrained in different schemes: supervised (called \supViT for brevity) and unsupervised (contrastive - \MoCo~\cite{moco} and autoencoder-based - \MAE~\cite{mae}). 
We applied end-to-end finetuning using the~whole PCam training set. As the pretrained transformer-based models operate on an input size of 224x224 and the images within PCam dataset are of size 96x96, we applied resizing (as it is done in~\cite{vpt}). We did not apply any standarization to keep the same preprocessing procedure as in the convolutional models. The hyperparameters such as base learning rate (\textit{lr\textsubscript{base}}) and weight decay (\textit{wd}) were the~same as the~ones used by the authors of~\cite{vpt} (details in Table~\ref{tab:params}). Compared to the paper, we decreased the batch size from 128 to 64 and shortened the training procedure. The~base learning rate was linearly scaled using the formula $lr=lr_{base}*batch\_size/256$~\cite{scaling}. The~models were finetuned for 5 epochs with Adam optimizer and with a linear learning rate scheduler increasing to the~final $lr$ value. We took the model from the epoch that resulted in the biggest accuracy on the validation set. The fact that a small number of epochs was enough to achieve satisfactory accuracy was due to the big size of the training set (262,144 samples).

\begin{table}[ht!]
\caption{Hyperparameters used for finetuning of transformer-based models on PCam dataset.}
\begin{tabular}{ccc}
\label{tab:params}
\textbf{model} & \textit{\textbf{lr\textsubscript{base}}} & \textit{\textbf{wd}} \\ \hline
\Swin & 0.0001 & 0.001 \\
\supViT & 0.001 & 0.01 \\
\MAE & 0.0001 & 0.0001 \\
\MoCo & 0.0001 & 0.0001
\end{tabular}
\end{table}

\section{Results}
In the PCam dataset, there are duplicates of images due to the~procedure used to create it. In all experiments, we use the test set after the removal of the duplicates. First, we perform the inference using a test set with original images (with full context size) using the analysed models. As a result, the reference values of performance metrics are obtained (Table~\ref{tab:metrics}).

\begin{table}[ht!]
\caption{Deep Learning models' performance when full available contextual information is given (so-called reference performance).}
\footnotesize
\begin{tabular}{ccccc}
\label{tab:metrics}
\textbf{model} & \textit{\textbf{acc\textsubscript{ref}}} & \multicolumn{1}{l}{\textit{\textbf{precision\textsubscript{ref}}}} & \textit{\textbf{recall\textsubscript{ref}}} & \multicolumn{1}{l}{\textit{\textbf{AUC\textsubscript{ref}}}} \\ \hline
\ResNet18 & 0.8786 & 0.9396  & 0.7754 &  0.9477\\
\DenseNet121 & 0.8941 & 0.9498 & 0.8029 & 0.9904 \\
\Swin & 0.9172  & 0.9680 & 0.8405 & 0.9716 \\
\supViT & 0.9160 & 0.9514 & 0.8537 & 0.9734  \\
\MAE & 0.9173 & 0.9513 & 0.8568 & 0.9746\\
\MoCo & 0.9143  & 0.9617 & 0.8397 & 0.9708
\end{tabular}
\end{table}
\normalsize

The analysed models have similar reference performance. However, it is observed that transformer-based models perform slightly better than convolutional models. Note that the convolutional models have smaller capacity as they have much less tunable parameters in millions (\ResNet18 - 11.7, \DenseNet121 - 8) than transformer-based models (\Swin~- 86.7, \textit{ViT} - 85.8), which can be a~source of the~difference. 


\subsection{Image dimensions mapping}
As already mentioned transformer-based models operate on the resized PCam images (224x224) whereas the~convolutional models - on the original size of the images (96x96). Therefore, the maximum size of the context and the central square is increased. This mismatch is important when analysing the results on a common axis. Note that the mapping from the size of 96x96 to 224x224 is of factor 2.33 which makes the alignment difficult when trying to map the values of context size in the two scenarios. As a solution, we decrease the context size in the~resized images by 7 pixels as opposed to the scenario with images of original size where it is done pixel by pixel. Cutting out the context by every 7\textsuperscript{th} pixel in the resized image translates to cutting every 3\textsuperscript{rd} pixel in the original image. As a consequence in the plots in the following sections, there will be more points on the curves referring to the~convolutional models than the transformer-based ones.
Note that the context size of 32 within the original images maps to the value of 74.67 within the resized images, therefore, between the 74\textsuperscript{th} and 75\textsuperscript{th} pixel (which we additionally analyse despite the policy to cut off the context by every 7\textsuperscript{th} pixel). To aggregate the results from the two context sizes, we apply a conservative approach. 
We take the~score that is the furthest from the~label meaning that (1)~if the~correct class is 1, we take the lowest probability from the~two pixels (74\textsuperscript{th} and 75\textsuperscript{th}), (2) if the~label is 0, we take the biggest probability.

\subsection{Drop of performance with the~decrease of context size}
\begin{figure*}[htb!]
\centering
\includegraphics[width=1\textwidth]{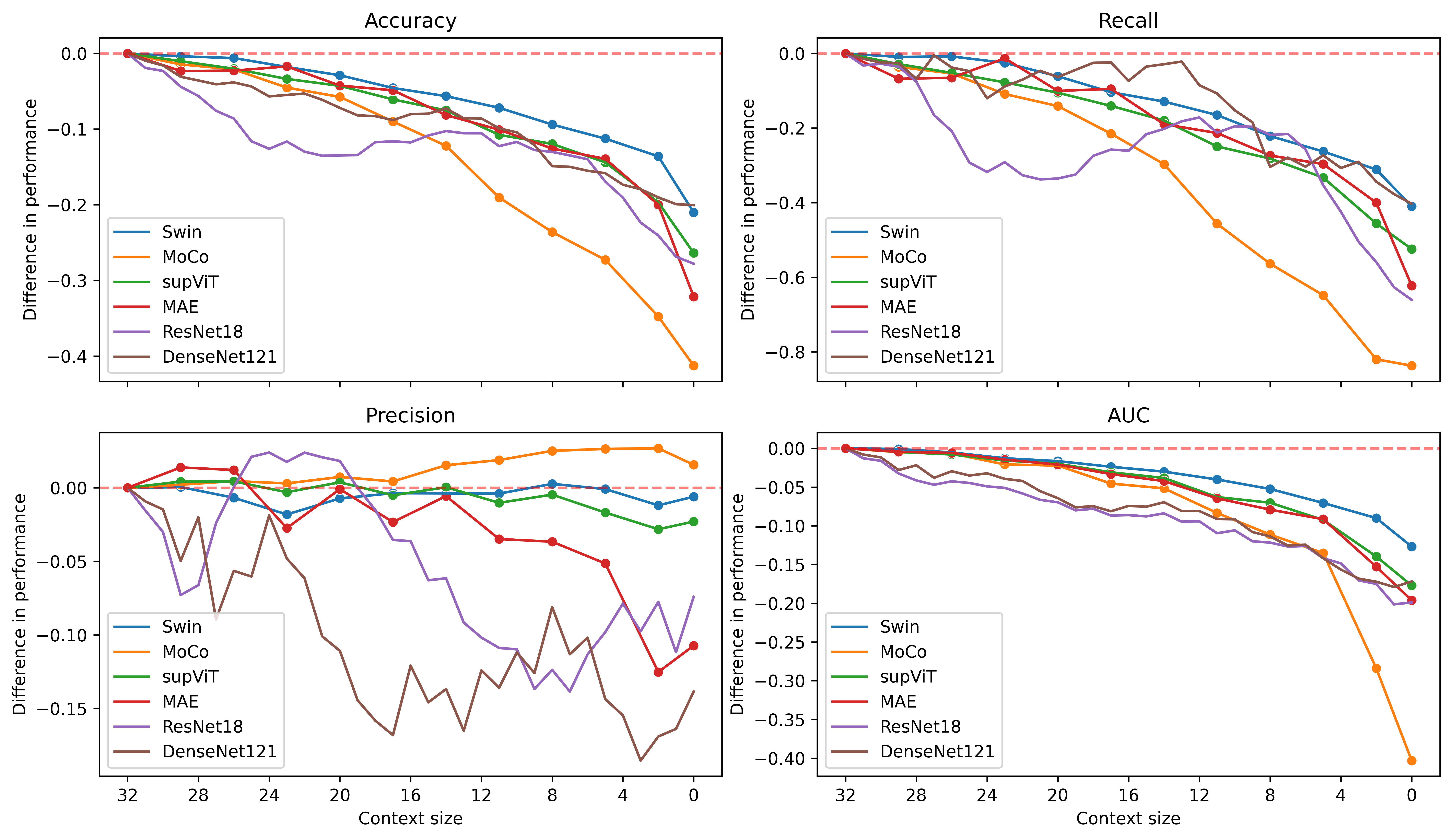}
\caption{The performance gap when the context size is limited. The performance metrics of Deep Learning models decreased by the respective reference values (when full context is available) under different context sizes are shown. Note that the values on the \textit{x}-axis are in decreasing order which makes an interpretation of the~experiments easier. The \textit{y}-axis is not shared within the subplots so that the variations of results for different models are more visible. The markers on the curves corresponding to transformer-based models highlight a smaller number of data points than in the case of convolutional models.}
\label{fig:performance_drop}
\end{figure*}

Having reference values of performance metrics, we limit the context size within the input images by applying a~black border to hide parts of the contextual information. The~performance metrics for different models are shown in Figure~\ref{fig:performance_drop}.

It is observed that the metric that is the most affected by limiting the amount of contextual information available at the prediction time is \textit{recall}. This metric is especially important in healthcare where false negatives are of the greatest concern. The biggest drop in \textit{recall} (0.84) occurs in the case of \MoCo model which starts to predict mostly one class - normal. At the~same time, in the case of this model, the \textit{precision} curve slightly increases with the decrease of context size (up to 0.016 when $context\_size=0$) even though the reference value was already very high (0.9617). The two models that experience the~smallest drop of \textit{recall} are \Swin (0.41) and \DenseNet121 (0.40).

Note that \textit{precision} seems to be the least impacted metric by the limitation of contextual information. The~most considerable decrease is observed in the~case of \DenseNet121 (0.14) and \MAE (0.11), however, in the~first case, the~significant drop is observed even when the~amount of context is only slightly limited. 

In the case of \textit{accuracy} and \textit{AUC} plots, all the models behave similarly except \MoCo which has the biggest drop of about 0.41. Interestingly, the \textit{accuracy}, \textit{recall} and \textit{precision} curves of \ResNet18 experience significant fluctuation over different context sizes. In the case of \textit{accuracy} and \textit{recall}, local minimum is observed for a~context size of about 24, whereas the local maximum~at~8. 
\newline

Note that by the fact that fewer data points are depicted in the case of transformer-based models than in the~convolutional models, the smoothness of the curves cannot be compared between the two families of models, unlike the general trends. Overall, it seems that the analysed convolutional models are more sensitive to the lack of contextual information than the transformer-based models (except \MoCo). However, the differences do not seem significant taking into account the gap in the models' capacity.


\subsection{Consistency of predictions over different context sizes}
Despite the changes in models' performance, the predictions for some images are the same over different context sizes. It turns out that the model that generates the consistent predictions most often is \Swin and at the second place - \supViT (see Table~\ref{tab:sensitive}). The rest of the models are far behind when it comes to the percentages of images with agreeing predictions. The most sensitive to the decrease of the context size is \DenseNet121.

\begin{table}[h!]
\caption{Percentage of images that get the same prediction regardless of the context size.}
\label{tab:sensitive}
\footnotesize
\begin{tabular}{ccccccc}
\ResNet18 & \DenseNet121 & \Swin & \supViT & \MAE & \MoCo \\ \hline
 54.66  & 52.39 & 72.41 & 69.06 & 60.76 & 58.20
\end{tabular}
\end{table}
\normalsize

\subsection{Misleading pieces of context}
\begin{figure}[htb!]
\centering
\includegraphics[width=1\linewidth]{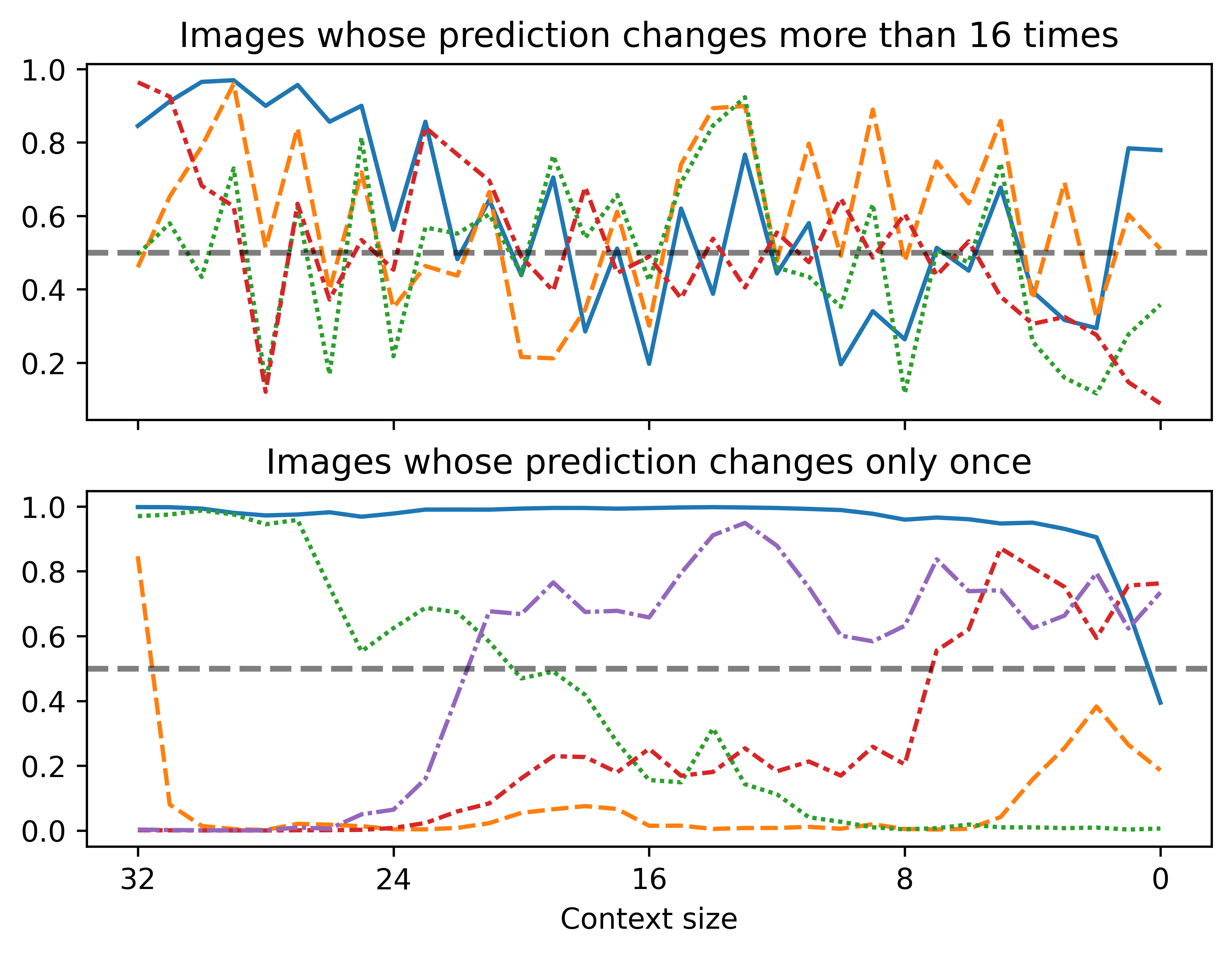}
\caption{The misleading nature of context depending on its size (on the example of \DenseNet121). The probability of the class tumour is shown in two cases: all images that experienced the change in the prediction in more than half of all context sizes (top), and a sample of images that changed the~class only once (bottom).}
\label{fig:misleading}
\end{figure}

The decrease in performance results means that the models change their predictions for some images depending on the amount of contextual information. We distinguish two types of images (1) the~ones that undergo the~change of prediction given a particular model only once (2) the~ones that experience the change of prediction more than once (referred to as \textit{'swinging images'}). We present how the~probability of class tumour changes depending on the context size for the sample images from the aforementioned two types in the case of \DenseNet121 (Figure~\ref{fig:misleading}). It is seen that the images that undergo frequent changes of the predictions are the ones that get the probabilities close to the threshold of 0.5. However, note that there is one sample (depicted in blue colour) that the model was initially confident about (probability above 0.8) but with the decrease of the context size, the~probability went down and started to oscillate around the threshold. It can be seen that in the case of images that experience the change of prediction only once, it~happens for different context sizes. 

\begin{figure}[h!]
\centering
\includegraphics[width=1\linewidth]{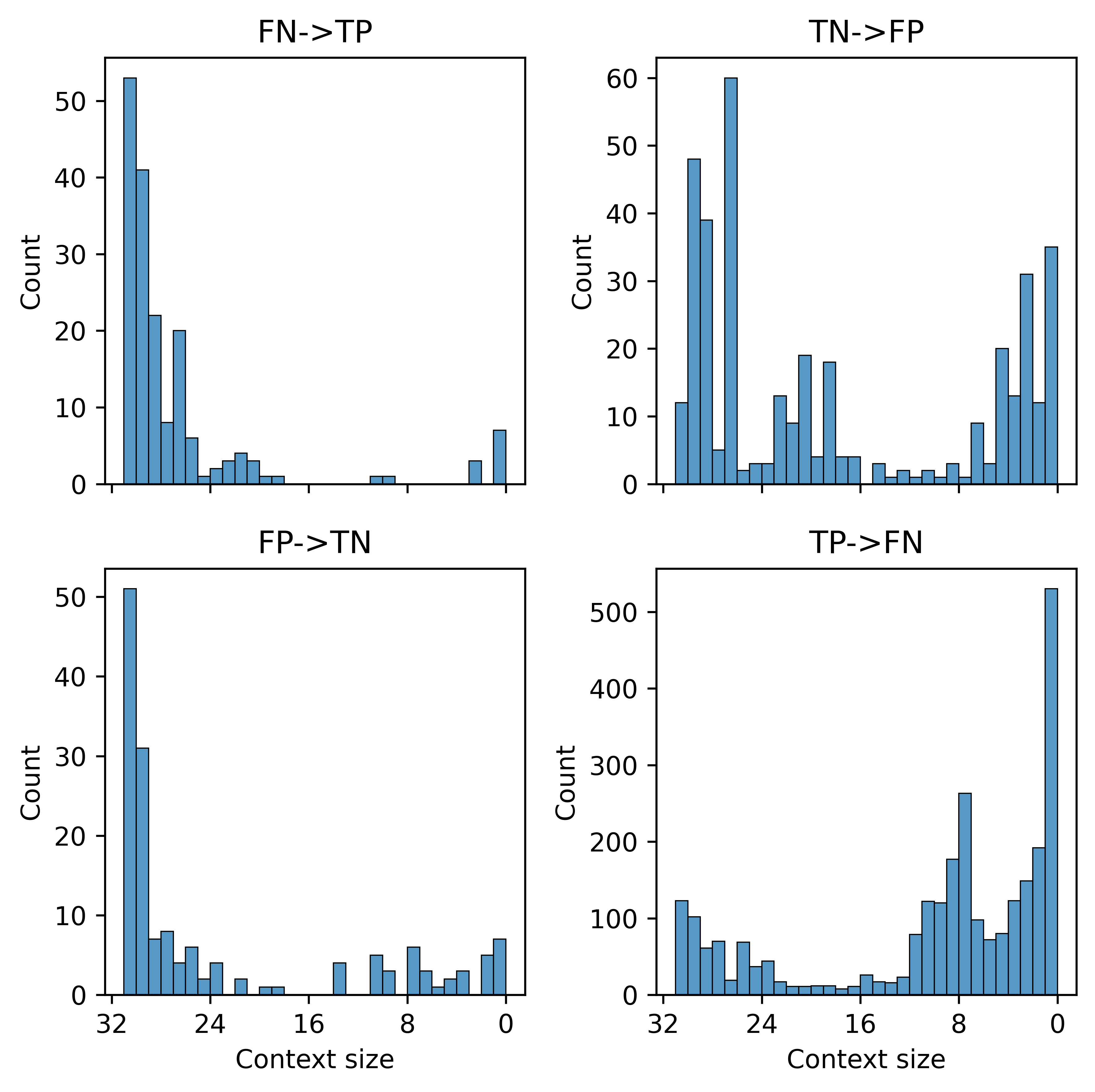}
\caption{Number of images undergoing the change of prediction for the first time given the particular context size with the distinction on the initial and consecutive model prediction (\textit{`swinging images'} not included). Note that for better visibility, the \textit{y}-axis is not shared between subplots. The bin width is equal to 1. The results are provided for the \DenseNet121.}
\label{fig:misleading_histogram}
\end{figure}

The analysis of when (at what context size) the model changes predictions was performed. We analyse the number of images undergoing the particular direction of a~change for the first time at the given context size. In the analysis, the \textit{`swinging images'} are not included. We show results only for \DenseNet121 as for illustration (Figure~\ref{fig:misleading_histogram}). Interestingly, it can be seen that the changes of predictions occur mostly for the extreme values of context size - either the small or the big ones. This observation holds for other models except \ResNet18 where the shift TN->FP happens mostly for moderate values of context size, however, the scale of the shift is small as for the given moderate context size there are a maximum of 11 images that undergo a particular change of prediction. 

\subsubsection{Possible interpretation}
The shift FP->TN could suggest that the model has seen some tumour cells in the context area as it predicted the~class tumour but by limiting the amount of contextual information, we cut off the pixels containing these tumour cells and the model outputs the correct class `normal'. Such behaviour of the model could be understandable as the model was not explicitly said to base its opinion only on the central square. \\
The other shifts are more difficult to find a reasonable explanation of. 
For example, the shift FN->TP could mean that initially, the model focused too much on contextual information where there was only normal tissue. When this distraction in the form of contextual information was taken away, the model paid only attention to the key central part and spotted the tumour cells and as a~consequence output the correct class. However, such a~behaviour of the model is not desirable. \\
In the aforementioned possible interpretations, we focused on the ones that do not require domain knowledge. However, the tissue structure and some spatial relationships that were partially covered by the black borders may potentially also play a role in the changes of model predictions. 

\subsubsection{Summarized model behaviour}
\begin{figure}[hbt!]
\centering
\includegraphics[width=1\linewidth]{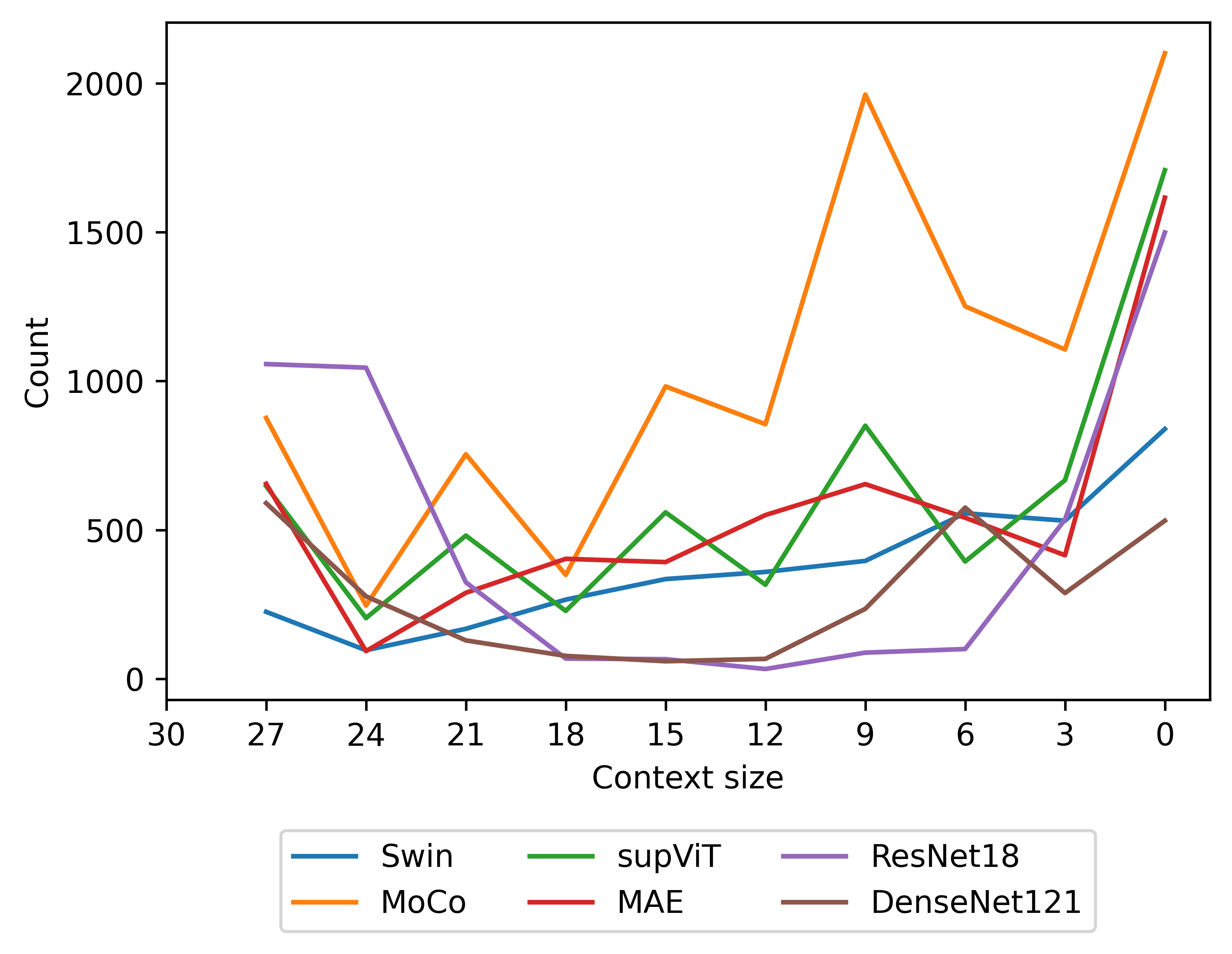}
\caption{Number of images undergoing the change of prediction for the first time given the particular context size (\textit{`swinging images'} not included). To account for the fact that in the~case of transformer-based models, we cut the context size by every 3 pixels (after mapping the image size back to the original one) not by 1 pixel as in the case of convolutional models (see Section \textit{Image dimensions mapping}), we summed the number of images undergoing the change from three consecutive context sizes. By the application of the aforementioned `normalization', the curves of convolutional and transformer-based models are on the same scale and have the same number of data points).}
\label{fig:misleading_number_all}
\end{figure}

The summarized results from the histograms (without the~distinction on the type of prediction shift) corresponding to different models are shown in Figure~\ref{fig:misleading_number_all}. 
It turns out that indeed (as shown in Figure~\ref{fig:misleading_histogram}) in the case of \DenseNet121, the most changes occur for extreme context sizes but it is even more visible in the case of \ResNet18. In \supViT and \MAE, the biggest boost of the number of changes occurs when the context size is significantly reduced. The most changes in total are observed when \MoCo is used. 

\begin{figure}[hbt!]
\centering
\includegraphics[width=1\linewidth]{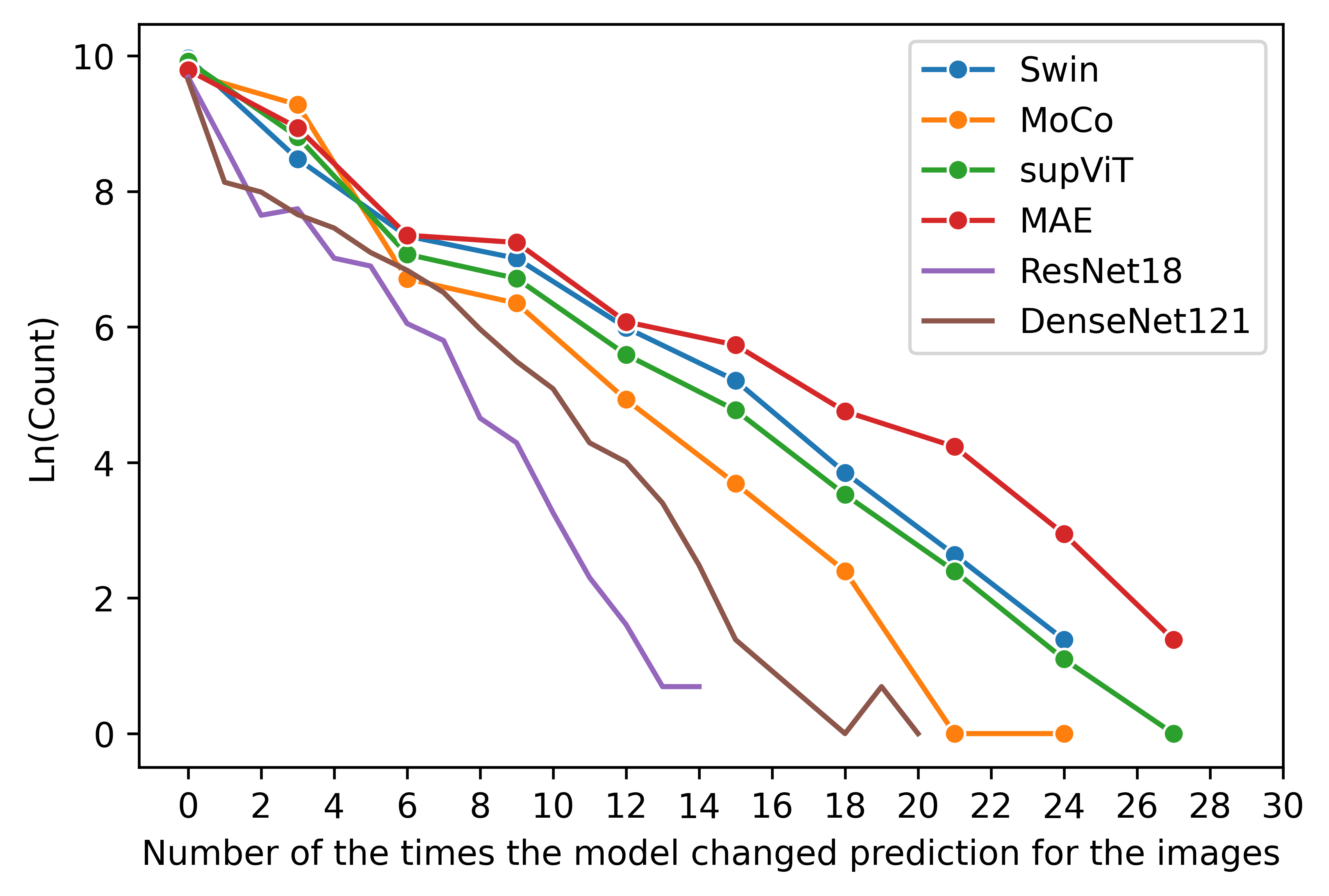}
\caption{The logarithm of the number of images that experience the particular number of prediction changes given a particular model. To account for the fact that the maximum possible number of prediction changes in the case of transformer-based models was 10 and in the case of convolutional models - 31, the values on $x$-axis are multiplied by the~factor of 3 (for transformer-based models) to put them on the same scale as convolutional models allowing a fair comparison. 
} 
\label{fig:number_all_changes}
\end{figure}

Lastly, it was analysed how many images in total experience a particular number of prediction changes given a~particular model (Figure~\ref{fig:number_all_changes}). It turns out that the models that change predictions more than 24 times per image (out of 30 analysed possibilities, after rescaling explained in the Figure's caption) are \MAE and \supViT. However, these are very rare - in the case of \MAE, there are only four such images, and in the case of \supViT - only one. Note that the area under the curves cannot be compared between the models from the convolutional and transformer families as in the~latter there are fewer data points. However, when comparing \DenseNet121 and \ResNet18, it is visible that for the \textit{`most swinging'} images, the changes of predictions are more frequent in the case of \DenseNet121 than \ResNet18. The four \textit{`most swinging'} images given \DenseNet121 are shown in Figure~\ref{fig:misleading}. The most similar models in behaviour are \supViT and \Swin where the relationship between the logarithm of the number of images undergoing the particular number of changes and the~number of changes is almost linear.

\subsubsection{Models' agreement}
It is analysed whether the same images are confusing to the models when the context size is limited. We investigate how many \textit{`swinging images'} are in common for any pair of models regardless of context size when the changes of predictions occur. We analyse transformer-based models and convolutional ones separately. It turns out that the biggest agreement between the transformer-based models is in the case of \Swin and \MAE (1176 cases) whereas the smallest agreement is between \supViT and \MoCo (502 cases). In the case of \ResNet18 and \DenseNet121, there are 4822 \textit{`swinging images'} in common, therefore, they seem much more alike than transformer-based models even though three out of four analysed transformer-based models have the same architecture (ViT) but differ in pretraining scheme.

\section{Conclusions}
In the work, we investigate whether the Deep Learning models for vision are sensitive to contextual information when making predictions on histopathological data. It turns out that when the context size is limited, the models achieve worse performance than in the case when full context is available which means that context is important at prediction time. It is observed that depending on the amount of contextual information, the model can output different predictions for a given image. We evaluate the behaviour of models that have similar reference performance metrics (when full access to a context is provided) in the case when the size of contextual information is decreased. It turns out that the model that is the most sensitive to the limitation of context size is \MoCo. It may possibly be attributed to the fact that the model was pretrained in a contrastive way but it requires further investigation. We observe that there are images of two types - the ones that undergo one change of prediction and the \textit{`swinging images'}. For the latter, it would be interesting to consult the images with a histopathologist to verify whether indeed in these cases, the context may be misleading. Moreover, in the future, the possible interpretations of the obtained results could be complemented by the analysis with the use of heatmap-based XAI techniques.

\begin{acknowledgments}
The project was funded by funded by the Warsaw University of Technology within the Excellence Initiative: Research University (IDUB) programme. The work was carried out with the support of the Laboratory of Bioinformatics and Computational Genomics and the High Performance Computing Center of the Faculty of Mathematics and Information Science, Warsaw University of Technology.
\end{acknowledgments}

\Urlmuskip=0mu plus 1mu\relax

\bibliography{sample-ceur}

\appendix

\end{document}